\documentclass[letterpaper, 10 pt, conference]{ieeeconf} 
\IEEEoverridecommandlockouts

\date{October 2024}

\newcommand{\shnote}[1]%
    {\textcolor{magenta}{ #1}}

\newcommand{\zgnote}[1]%
    { \textbf{\textcolor{red}{#1}} }

\newcommand{\tinit}{t}
\newcommand{\xinit}{x}
\newcommand{\treach}{t_r}
\newcommand{\tavoid}{t_a}

\newcommand{\tvar}{s}
\newcommand{\tstep}{\delta }

\newcommand{\maxlevel}{ \mathcal I_M}
\newcommand{\thor}{0} % Time horizon
\newcommand{\state}{x} % Tracker state

\newcommand{\dyn}{f} % Tracker dynamics
\newcommand{\cdyn}{g_\ctrl} % control dynamics
\newcommand{\ddyn}{g_\dstb} % control dynamics

\newcommand{\ctrl}{u} % Tracker control
\newcommand{\dstb}{d} % Disturbance
\newcommand{\csig}{\ctrl(\cdot)} % Tracker control
\newcommand{\dsig}{\dstb(\cdot)} % Disturbance
 
\newcommand{\dset}{\mathcal{D}}
\newcommand{\dfset}{\mathbb{D}}
\newcommand{\dmap}{\lambda} % dstb strategy
\newcommand{\Dmap}{\Lambda} % dstb strategy set

\newcommand{\loss}{c}
\newcommand{\reward}{r}

\newcommand{\acs}{\mathcal{U}_a}

\newcommand{\racs}{\mathcal{U}_{\mathcal G}^r}
\newcommand{\aacs}{\mathcal{U}_{\mathcal C}^a}
\newcommand{\aacsinf}{\mathcal{U}_{\mathcal C}^{a,\infty}}
\newcommand{\sacs}{\mathcal{U}_{p}^s}

\newcommand{\racsd}{\mathcal{U}_{\mathcal G,d}^r}
\newcommand{\aacsd}{\mathcal{U}_{\mathcal C,d}^a}

\newcommand{\sacsd}{\mathcal{U}_{p,d}^s}

\newcommand{\cset}{\mathcal{U}} %tracker control set
\newcommand{\cfset}{\mathbb{U}} %tracker control funciton set
\newcommand{\traj}{\xi(\tvar;\tinit, \xinit,\csig,\dsig)}
\newcommand{\trajmap}{\xi(\tvar;\tinit, \xinit,\csig,\dmap)}

\newcommand{\mrcis}{\mathcal{I}_m}

\newcommand{\dom}{\mathcal{D}_\gamma}

\newcommand{\goal}{\mathcal{R}}

\newcommand{\obs}{\mathcal{C}}

\newcommand{\valfunc}{V} % Value function
\newcommand{\clvf}{\valfunc_\gamma^\infty} % CLVF 
\newcommand{\Vr}{\valfunc_r} % 
\newcommand{\Va}{\valfunc_a} % 
\newtheorem{remark}{Remark}
\newtheorem{proposition}{Proposition}
\newtheorem{definition}{Definition}

\newtheorem{theorem}{Theorem}
\newtheorem{lemma}[theorem]{Lemma}

\usepackage{graphicx}
\usepackage{booktabs}
\usepackage{amsmath} % assumes amsmath package installed
\let\labelindent\relax
\usepackage{enumitem}

\usepackage{amsfonts}
\usepackage{amssymb}  % assumes amsmath package installed
\usepackage{mathtools} %% for \coloneqq
\usepackage{mathrsfs}
\usepackage{balance}
\usepackage{url}
\usepackage{hyperref}
\usepackage{algorithm}
\usepackage{algpseudocode}
\DeclareMathAlphabet{\mathpzc}{OT1}{pzc}{m}{it}
\usepackage{makecell}
\usepackage[noadjust]{cite}
\usepackage{xcolor}
\usepackage{comment}
\begin{document}

\title{Reach-Avoid-Stabilize Using Admissible Control Sets}

\author{Zheng Gong*, \IEEEmembership{Student Member, IEEE}, Boyang Li*, and Sylvia Herbert, \IEEEmembership{Member, IEEE}
\thanks{This work is supported by ONR YIP (\#N00014-22-1-2292) and the UCSD JSOE Early Career Faculty Award.}
\thanks{*Both authors contributed equally to this work. All authors are in Mechanical and Aerospace Engineering at UC San Diego (e-mail: 
\{{\href{mailto:zhgong@ucsd.edu}{zhgong}, \href{mailto:bol025@ucsd.edu}{bol025}, \href{mailto:sherbert@ucsd.edu}{sherbert}\}@ucsd.edu.})
}}

\maketitle

\begin{abstract}
Hamilton-Jacobi Reachability (HJR) analysis has been successfully used in many robotics and control tasks, and is especially effective in computing reach-avoid sets and control laws that enable an agent to reach a goal while satisfying state constraints. However, the original HJR formulation provides no guarantees of safety after a) the prescribed time horizon, or b) goal satisfaction. The reach-avoid-stabilize (RAS) problem has therefore gained a lot of focus: find the set of initial states (the RAS set), such that the trajectory can reach the target, and stabilize to some point of interest (POI) while avoiding obstacles. Solving RAS problems using HJR usually requires defining a new value function, whose zero sub-level set is the RAS set. The existing methods do not consider the problem when there are a series of targets to reach and/or obstacles to avoid. We propose a method that uses the idea of admissible control sets; we guarantee that the system will reach each target while avoiding obstacles as prescribed by the given time series. Moreover, we guarantee that the trajectory ultimately stabilizes to the POI. The proposed method provides an under-approximation of the RAS set, guaranteeing safety. Numerical examples are provided to validate the theory. 
\end{abstract}
%%%%%%%%%%%%%%%%%%%%%%%%%%%%%%%%%%%%%%%%%%%%%%%%%%%%%%%%%%%%%%%%%%%%%%%%%%%%%%%%%%%%%%%%%%%%%%%%%%%%%%%%%%%%%%%%%%%%%%%%%%%%%%%%%%%%%%%%%%%%%%%%%%%%%%%%%%%%

\section{Introduction}

% \zgnote{Para1: First introduce the general importance of safety and liveness control. Then introduce CBF/CLF/HJ/CLVF}
Hamilton-Jacobi Reachability (HJR) analysis is a powerful tool for solving  \textit{liveness} (i.e., goal-satisfaction) and \textit{safety} problems. It defines a value function, whose zero sublevel set is the set of initial states that can reach a target set and/or avoid an obstacle set given a prespecified time horizon~\cite{fisac2015reach,bansal2017hamilton}. The gradients of this value function inform the optimal control policy over the time horizon; for reach-avoid problems, this policy would reach the goal in minimum time while maintaining state constraints. The value function is computed by solving the Hamilton-Jacobi-Issacs variational inequality (HJI-VI) via dynamic programming (DP) in the prescribed time horizon. %It also serves as a verification tool for optimal controls derived from other approaches~\cite{bardi2008optimal}. 
In addition to the `curse of dimensionality' from DP, another major drawback of standard formulations for HJR is that stabilizability (which is in infinite-time horizon) is not considered, and the system may leave the goal and violate safety after a) initially reaching the goal, or b) the prescribed time horizon.  

To mitigate this issue, the reach-avoid-stay problem is considered and studied. One natural way is to combine Control Lyapunov Functions (CLF)~\cite{freeman1996control,sontag1999control,primbs1999nonlinear} and Control Barrier Functions (CBF)~\cite{xu2015robustness,ames2016control,ames2019control}. These functions have been intensively studied during the past decades. Once found, their gradients can be used to derive constraints on the control, and combined through an optimization problem~\cite{garg2019control,zeng2021safety}. However, a feasible solution is generally not guaranteed for this optimization problem. The Control Lyapunov-barrier function~\cite{romdlony2016stabilization,braun2020comment,meng2023lyapunov} and several of its extensions are proposed, with the idea of stabilizing the system to the origin while staying inside the control invariant set. The main drawbacks are the lack of a constructive way to find those functions (relying instead on data-driven approximations or hand-designed approaches)~\cite{dawson2022safe}. Works that solve the reach-avoid-stay problem from the HJR perspective have also been considered~\cite{chenevert2024,li2024certifiable}; these works usually require defining new value functions and using DP to solve them.

%One common limitation of the above-mentioned methods is that they do not consider multiple-target problems. 

In this paper, we consider the reach-avoid-stabilize (RAS) problem. Moreover, assume that we are given a set of targets and obstacles, together with two time series, specifying the time horizon we need to reach each target and avoid each obstacle. After reaching the last target, we are asked to stabilize to a point of interest (POI) while avoiding the obstacles (if any). To the best of our knowledge, no HJR-based method exists to solve this RAS problem. 

The proposed method stems from HJR analysis but does not require defining any new value functions. The concept of admissible control set (ACS), which is used to solve the `curse of dimensionality' of HJR analysis~\cite{he2023efficient,gong2024synthesizing}, is the key component of our method. The ACS is obtained from the HJR value function and can be viewed as state-and-time-wise constraints on the control inputs that render the system satisfying the \textit{liveness} or \textit{safety} requirement. Once the ACSs for each target and obtscale are found, a forward propagation approach is applied to find the set of initial states that satisfy the RAS requirement.   %The ACS has also been studied from the perspective of smoothing the `least-restrictive control'~\cite {borquez2024safety}. Also,~\cite{lavanakul2024safety} uses learning-based approaches that bypass the need to find a value function to derive the control. 

The paper is organized as follows: Section~\ref{SEC2} provides background information and formally defines the RAS problem. Section~\ref{SEC3} provides a detailed discussion on the ACS in the case of stabilization, and a quick review of the ACS for reach and avoid problems. Section~\ref{SEC4} shows how the `admissible control signal' can be synthesized from the ACS in a mathematically rigorous way. Section~\ref{SEC5} provides algorithms to under-approximate the RAS set. 
%and generalizes to the case where there are multiple targets (obstacles) to be reached (avoided) in different time horizons. 
Numerical examples are provided in Section~\ref{SEC6}, which validates the theory. The main contributions are: 
\begin{enumerate}
    \item We define the set of stabilizing control inputs %$\sacs (\state)$ 
    for general nonlinear control-affine systems. We provide a detailed discussion on guaranteeing exponential stabilizability and asymptotic stabilizability. 
    % \item We show that the ASC can be tuned, without recomputing the value functions. Further, when the current disturbance is known, the ACS can be accordingly changed for different purposes. 
    \item Given a set of targets (obstacles) and a corresponding time series that specifies the time horizon to reach each target (avoid each obstacle),  we propose a computationally efficient algorithm to under-approximate the RAS set. Further, the order of reaching each target within the corresponding time horizon can be switched, and the same for the obstacles.  
    % \item We further extend the algorithm to solve the problem with multiple targets and obstacles by decoupling the time horizon. More specifically, given a set of targets (or obstacles) and a time series specifying the time horizon you want to reach the targets (or avoid the obstacles), our algorithm can find the RAS set in this setting. Further, the order or reaching different targets can be easily switched.
    \item Numerical examples are provided to validate the theory and show computational efficiency. 

    % \item More general decomposition results are provided to tackle the curse of dimensionality.
\end{enumerate}

%%%%%%%%%%%%%%%%%%%%%%%%%%%%%%%%%%%%%%%%%%%%%%%%%%%%%%%%%%%%%%%%%%%%%%%%%%%%%%%%%%%%%%%%%%%%%%%%%%%%%%%%%%%%%%%%%%%%%%%%%%%%%%%%%%%%%%%%%%%%%%%%%%%%%%%%%%%%
\section{Background} \label{SEC2} 
Consider the control and disturbance-affine system
\begin{align} \label{eqn:dynsys}
    \frac{d \state}{d \tvar} = \dot \state = \dyn (\state) + \cdyn (\state) \ctrl + \ddyn(\state)  \dstb,
\end{align}
where $t< 0$ is the initial time, $\state(\tinit) = \xinit $ is the initial state, $\tvar \in [\tinit, \infty)$ is the time variable, $\state \in \mathbb R^n$ is the state, $\ctrl \in \cset$ and $\dstb \in \dset$ are the control and disturbance input respectively. Assume the dynamics $\dyn$, $\cdyn$, $\ddyn$ are Lipschitz continuous in $\state$, and the control signal $\csig$ and disturbance signal $\dsig$ are drawn from the set of measurable functions that map the time to $\cset$ and $\dset$ respectively: 
\begin{align*}
    &\cfset:= \{ \csig: [\tinit, \infty) \mapsto \cset, \csig \text{ is measurable}  \}, \\
    & \dfset:= \{ \dsig: [\tinit, \infty) \mapsto \dset, \dsig \text{ is measurable}  \}.
\end{align*}
Under these assumptions, a unique trajectory can be obtained given initial condition $\xinit$, denoted as $\traj$, and in short $\xi(\tvar)$. We further assume the disturbance signal can be determined as a non-anticipative strategy $\dmap \coloneqq \cfset \mapsto \dfset$ with respect to the control signal, which gives an instantaneous advantage to the disturbance. The set of non-anticipative strategies is denoted as $\Dmap$, i.e., $\dmap \in \Dmap$. 

\begin{definition}[RAS Set]  \label{Def: RAS_set}  Given $q_1$ targets $\{\goal_i\}_{i = 1}^{q_1}$ and $q_2+1$ obstacles $\{\obs_i\}_{i = 1}^{q_2}$, $\obs$, with reach time series $T_{r} = [t_{r,0}, t_{r,1}, ... , t_{r,q_1}]$, $t_{r,0} = t$, avoid time series $T_{a} = [t_{a,0}, t_{a,1}, ... , t_{a,q_2}]$, $t_{a,0} = t$, and a POI $p$, the RAS set is 
    \begin{align*}
    \mathcal {RAS} & (T_r,T_a , p) = \{ \xinit:  \exists \csig, \forall \dsig  \text{ s.t. } \\ 
    & \exists \tvar \in [t_{r,i-1}, t_{r,i}], \traj \in \goal_i \hspace{.5em} \forall i = 1...q_1,  \\ 
   \text { and } & \forall \tvar \in [t_{a,j-1}, t_{a,j}], \traj \notin \obs_j \hspace{.5em} \forall j = 1...q_2 \\ 
    \text { and } & \forall \tvar \in [\tinit, \infty), \traj \notin \obs_\infty \\
    \text { and } &  
    \lim_{\tvar \rightarrow \infty } \min_{a \in \partial \mrcis(p)} \| \traj - a \| = 0 \}.
\end{align*}
\end{definition}

We assume the targets and obstacles can be charaterizezd as zero sub-level sets of the target functions $\{ \reward _i \}_{i = 1}^{q_1}$, and zero super-level sets of the obstacle functions $\{ \loss _i \}_{i = 1}^{q_2}$ and $\loss$. 

% Given a POI $p$, $q_1$ targets (i.e. goals) $\{\goal_i\}_{i = 1}^{q_1}$ and $q_2+1$ obstacles $\{\obs_i\}_{i = 1}^{q_2}$ and $\obs$, which are given by the zero sub-level (super-level) sets of the target and obstacle functions $\{ \reward _i \}_{i = 1}^{q_1}$, $\{ \loss _i \}_{i = 1}^{q_2}$ and $\loss$. Consider two time series: one for reach times $T_{r} = [t_{r,0}, t_{r,1}, ... , t_{r,q_1}]$ and one for avoid times $T_{a} = [t_{a,0}, t_{a,1}, ... , t_{a,q_2}]$, with $t= t_{r,0} <t_{r,1}< ... < t_{r,q_1}$ and $t= t_{a,0} < t_{a,1}< ... < t_{a,q_2}$. 

In other words, the RAS set is the set of initial states s.t. for all $i = 0,...q_1, j=0,...,q_2$ the trajectory 1) reaches $\goal_i$ within $[t_{r,i-1}, t_{r,i}]$, 2) while avoiding $\obs_j$ for all $\tvar \in [t_{a,j-1}, t_{a,j}]$, 3) stabilize to a POI $p$, and 4) avoid $\obs$ for all time $\tvar \in [\tinit, \infty)$. %This set of initial conditions is called the RAS set. 

Note that we do not assume $p$ is a stable equilibrium point, therefore the best we can is to stabilize the system to the smallest robust control invariant set $\mrcis (p)$ as defined in~\cite{gong2024robust} (when $p$ is stable, $\mrcis(p) = p$). We further assume the region of stabilizability of $p$ has a non-empty intersection with the last target set $\goal_{q_1}$; otherwise, the RAS task has no solution.

\subsection{Hamilton-Jacobi Reachability}
A popular baseline approach for solving reach-avoid problems is HJR analysis. Consider only one target $\goal \coloneqq \{ \state: \reward(\state)\leq 0 \}$ and one obstacle $\obs \coloneqq \{ \state: \loss(\state)\geq 0 \}$.
% First, define the obstacle function $\loss$ whose zero superlevel set is the obstacle set: $\obs = \{ \state: \loss (\state) > 0 \}$ and the target function whose zero sublevel set is the target set: $\goal = \{ \state: \reward (\state) \leq 0 \}$. 
The RA set for $\tvar \in [\tinit , \thor]$ is defined as : 
\begin{align*} 
    \mathcal {RA} (\tinit) = \{ \xinit: & \forall \dmap \in \Dmap, \exists \csig \in \cfset  \text{ s.t. } \\ 
    & \exists \tvar \in [\tinit,\thor], \trajmap \in \goal,  \\ 
   \text { and } & \forall \tau \in [\tinit, \tvar], \xi (\tau;\tinit,\xinit, \csig, \dmap) \notin \obs \}.
\end{align*}
Here, we follow the convention in~\cite{bansal2017hamilton} and use $0$ as a terminal time horizon for the reach-avoid game. The value function of the reach-avoid problem is 
\begin{align*} % \label{eqn:RA_value}
    V(\state,\tinit) = \sup_{\dmap \in \Dmap} \inf_{\ctrl \in \cset} \min_{\tvar \in [\tinit,\thor]} \max \big\{ \reward (\xi(\tvar)),\max_{\tau \in [\tinit,\tvar] } \loss(\xi(\tau)) \big\} ,
\end{align*}
and the equation inside the $\sup \inf$ is the cost. The value function is the viscosity solution of the HJI-VI:
\begin{align} \label{eqn:HJIPDE}
    & 0 = \max \bigg\{  \loss(\state)-V(\state,\tinit) , \min \big\{ \reward(\state)-V(\state,\tinit) , \notag \\ &\frac{dV}{d\tinit} + \min_{\ctrl \in \cset} \max_{\dstb \in \dset} \frac{dV}{d\state} \cdot(\dyn(\state)+\cdyn(\state) \cdot \ctrl +\ddyn(\state) \dstb )
     \big\}  \bigg\}, 
\end{align}
with terminal condition
\begin{align*}
    V(\state,\thor) = \max \{ \reward(\state), \loss(\state) \}.
\end{align*}
Given a fixed $t$, the zero sublevel set of $V(\state,\tinit)$ is the RA set: it captures all the states that can reach $\goal$ at some time $s \in [\tinit,\thor]$, while avoid $\obs$ for all time in $[\tinit,s]$. 

% The reach and avoid value functions are given by 
% \begin{align}
%     \Vr(\state,\tinit) = \sup_{\dmap \in \Dmap} \inf_{\ctrl \in \cset} \min_{\tvar \in [\tinit,\thor]} \reward (\xi(\tvar))  , \label{eqn:Reach_value} \\ 
%     \Va(\state,\tinit) = \sup_{\dmap \in \Dmap} \inf_{\ctrl \in \cset}\max_{\tvar \in [\tinit,\thor] } \loss(\xi(\tau)),  \label{eqn:Avoid_value}
% \end{align}
% and are viscosity solutions of the HJIPDEs by removing the dependence on $\loss$ and $\reward$ in~\eqref{eqn:HJIPDE}. The zero sublevel (superlevel) set of $\Vr$ ($\Va$) is the reach (avoid) set, i.e., the set of initial states that can reach $\goal$ for some $\tvar \in [\tinit, \thor]$ (or avoid $\obs$ for all $\tvar \in [\tinit, \thor]$). 

% \begin{remark}
%     The RA set is different from simply finding the reach set and avoid set and taking the intersection of these two sets. The reason is that given a state, the control input for it to reach the goal could be different when it tries to avoid the obstacle. 
% \end{remark}

Once the value function is computed, the optimal control and worst case disturbance can be determined as 
\begin{align}
    \ctrl ^* (\state, \tvar) = \arg \min _{\ctrl} \frac{d V}{d\state}(\state,\tvar) \cdot \cdyn (x), \hspace{1em} \tvar \in [\tinit, \thor], \label{eqn:HJoptCtrl}\\
    \dstb ^* (\state, \tvar) = \arg \max _{\dstb} \frac{d V}{d\state}(\state,\tvar) \cdot \ddyn (x), \hspace{1em} \tvar \in [\tinit, \thor]. \label{eqn:HJoptDstb}
\end{align}
For better readability, we also write the reach and avoid value functions here:
\begin{align}
   & \Vr(\state ,\tinit) = \sup_{\dmap \in \Dmap} \inf _{\csig \in \cfset} \min_{\tvar \in [\tinit, \thor]} \reward (\xi(\tvar; \tinit, \state, \csig),\dmap), \label{eqn:reach_value} \\
   & \Va(\state ,\tinit) = \sup_{\dmap \in \Dmap} \inf _{\csig \in \cfset} \max_{\tvar \in [\tinit, \thor]} \loss (\xi(\tvar; \tinit, \xinit, \csig),\dmap). \label{eqn:avoid_value}    
\end{align}
The corresponding HJI-VI can be obtained by removing the dependence on $\loss$ and $\reward$ in~\eqref{eqn:HJIPDE}.

\begin{comment}
    \subsection{Control Lyapunov Function and Control Barrier Function} 
CBFs and CLFs are common approaches to solving the avoid problem and the stabilize problem respectively. 
\begin{definition} [Robust CLF]
    A continuous function $ V_{\text{clf}}: \mathbb R^n \mapsto \mathbb R$ is a local robust CLF for the equilibrium point $\state$ if, in a neighborhood $\mathcal O$ of $\state$, the following holds: (a) $V_{\text{clf}}$ is proper at $\state$, (b) $V_{\text{clf}}$ is positive definite on $\mathcal O$ and continuously differentiable on $\mathcal O$, and (c) for each $\state \in \mathcal O$, $\forall \dstb$, $\exists \ctrl \in \cset$ s.t.
    \begin{align*}
        \dot V_{\text{clf}} (\state) = \frac {dV_{\text{clf}}}{d\state} \cdot (\dyn(\state) +\cdyn(\state)  \ctrl + \ddyn(\state)  \dstb) < 0.
    \end{align*} 
\end{definition}

\begin{definition} [Robust CBF]
    Given a continuously differentiable function  $ V_{\text{cbf}}: \mathbb R^n \mapsto \mathbb R$, whose zero-superlevel set is characterized as $\mathcal B$. Then $ V_{\text{cbf}}$ is a robust CBF of \eqref{eqn:dynsys} if there exist a class $\mathcal K_\infty$ function $\alpha (\state)$, $\forall \state \in \mathcal B$, $\forall \dstb$, $\exists \ctrl$ s.t. 
    \begin{align*}
        \dot V_{\text{cbf}} (\state) = \frac {dV_{\text{cbf}}}{d\state} \cdot (\dyn(\state) +\cdyn(\state)\ctrl+\ddyn(\state)\dstb) \leq -\alpha(V_{\text{cbf}}).
    \end{align*}
\end{definition}
\end{comment}

\subsection{Control Lyapunov Value function}
As mentioned before, finding a CLF is challenging for nonlinear systems, especially when a disturbance exists and control is bounded. Hand-designed approaches are often used, which may lead to conservative approximations of the region of stabilizability. Therefore, we introduce the robust control Lyapunov function (R-CLVF)~\cite{gong2024robust}. The R-CLVF is an HJR-like function, which can be computed by solving a VI via dynamic programming until convergence. 

Given system~\eqref{eqn:dynsys} and $p$, the R-CLVF is: 
\begin{align}\label{eqn:CLVF}
    \clvf (\state ; p) = \lim_{\tinit \rightarrow -\infty} \sup _{\dmap \in \Dmap} \inf _{u\in \cset } \max _{\tvar \in [\tinit , \thor]} e^{\gamma(\tvar - \tinit)} h(\xi(\tvar) - p),
\end{align}
where $h(x) = \| x\| -a$, $\gamma > 0$ is a user-specified parameter that indicates the exponential decay rate, and $a$ is a constant depending on the system dynamics and $p$~\cite{gong2024robust}. The domain of the R-CLVF is denoted as $\dom$. When $p $ is a stable equilibrium point, $a=0$ and $0< \clvf (\state) < \infty$ for all $ \state \in \dom$ except for $p$, as $\clvf(p) = 0 $. Otherwise, $a> 0$ and we can find $\mrcis$ whose convex hall contains $p$, and $0< \clvf (\state) < \infty$ for all $ \state \in \dom \setminus \mrcis$. Further, the R-CLVF has been proven to be the viscosity solution to the following R-CLVF-VI
\begin{align}\label{eqn:CLVF-VI}
    0 = \max \biggl \{ h(\state) - \clvf(\state), &\min_{u \in \cset}\max_{\dstb \in \dset} \frac{d \clvf}{d \state} \cdot(\dyn(\state)+ \notag \\
    &\hspace{-2em} \cdyn(\state) \ctrl + \ddyn(\state) \dstb) + \gamma \clvf \biggr \}.
\end{align}
 From the R-CLVF-VI, there exists some $\ctrl \in \cset$ for all $\dstb \in \dset$ s.t. $\dot V_\gamma^\infty \leq -\gamma \clvf$. Therefore, $\dom$ is also called the region of exponential stabilizability~\cite{gong2024robust}. The optimal control $\ctrl^*$ and worst case disturbance $\dstb^*$ can be similarly obtained from~\eqref{eqn:HJoptCtrl}~\eqref{eqn:HJoptDstb}.

\section{ACS for Reach Avoid and Stabilizability} \label{SEC3}

Here we introduce the stabilize ACS and show a brief overview of the reach and avoid ACS. %For all three tasks, the HJR approach computes the value function with optimal control, i.e., for the reach/avoid case, the optimal control attempts to steer the system into the target as deep as possible  (or as far as possible to the obstacle), while for the reach/avoid problem of interest, simply reaching the zero level set of $\reward$ or avoiding the zero level set of $\loss$ is enough. Similarly, for the stabilize case, the trajectory tries to converge to $p$ as fast as possible, while an $\gamma$-exponential rate is enough. 

\subsection{Stabilize ACS}
One sufficient condition for the system trajectories to converge with $\gamma$-exponential rate to a POI $p$ is simply requiring $\dot V _\gamma^\infty < -\gamma \clvf$ for all states in $\dom$. Formally, define $\sacs: \mathbb R^n  \mapsto \cset$, 
\begin{align} \label{eqn:SCS_CLVF}
    \sacs (\state; \gamma) = \{ u: \frac{d\clvf }{d \state } \cdot \cdyn (\state) \ctrl  < -\gamma \clvf \notag \\ 
    - \frac{d\clvf}{d \state } \cdot (\dyn (\state) + \ddyn(\state)\dstb^* )\}  
\end{align}
as the stabilizing ACS. Here, $\gamma$ is the hyperparameter used to compute the R-CLVF. Note that since the inequality in~\eqref{eqn:SCS_CLVF} is strict, the stabilizing ACS can be empty.  
% \zgnote{need a better way to express the different $\gamma$ used, for CLVF and actual constraints in SCS. }

\begin{proposition} \label{prop:SCS_empty}
   $\sacs (\state; \gamma) =\emptyset$ if $\clvf (\state) > h(\state)$.

    \begin{proof}
       Assume $\clvf (\state) > h(\state)$. From~\eqref{eqn:CLVF-VI}, it must be: 
        \begin{align*}
            \min_{u \in \cset} \max_{\dstb \in \dset} \frac{d \clvf}{d \state} \cdot(\dyn(\state)+\cdyn(\state) \ctrl +\ddyn(\state) \dstb ) + \gamma \clvf = 0,
        \end{align*}
        which implies
        \begin{align*}
            \min_{u \in \cset} \frac{d\clvf }{d \state } \cdot \cdyn (\state) \ctrl  = -\gamma \clvf -  \frac{d\clvf}{d \state } \cdot (\dyn (\state) +\ddyn(\state) \dstb^* ),
        \end{align*}
        i.e., $\sacs (\state; \gamma) = \emptyset$.

        % Now, assume $\cset (\state) = u^*$ 
    \end{proof}
\end{proposition}

This can cause problems when synthesizing trajectories, because if $\sacs = \emptyset$, no control satisfies the exponential stabilizability condition with $\gamma$ rate. However, if we relax the condition in~\eqref{eqn:SCS_CLVF} with a smaller $\hat \gamma < \gamma$, this problem can be solved. Further, by choosing $\hat \gamma = 0$, we can guarantee asymptotic stabilizability, and by choosing $\hat \gamma > 0$, we can still guarantee exponential stabilizability with rate $\hat \gamma$. 

% This is not a problem, as the exponential stabilizability with any $\gamma>0$ is stronger than asymptotical stabilizability, and we could relax the constraints in~\eqref{eqn:SCS_CLVF} with a smaller $\gamma$ and still guarantee at least asymptotical stabilizability. 
% For practical implementation, this is also not a problem, and we present the following Proposition.

% This is a critical problem: if the SCS is empty, how are we supposed to synthesize the controller? This can be overcome by relaxing the $\gamma$ in~\eqref{eqn:SCS_CLVF} as stated in the following Proposition. 

\begin{proposition}
    Given any $0 \leq \hat \gamma < \gamma$, the stabilize ACS $\sacs (\state ; \gamma, \hat \gamma)$ is non-empty for all $\state \in \dom$, where 
    \begin{align}\label{eqn:SCS_CLVF2}
        \sacs (\state ; \gamma, \hat \gamma) = \{ u: \frac{d\clvf }{d \state } \cdot \cdyn (\state) \ctrl  < -\hat \gamma \clvf\notag \\ 
    - \frac{d\clvf}{d \state } \cdot (\dyn (\state) + \ddyn(\state)\dstb^* )\}.
    \end{align}
    
    \begin{proof}
        We only need to prove for the case when $\clvf(\state) > h(\state)$ because of~\eqref{eqn:CLVF-VI}. It follows directly from Prop.~\ref{prop:SCS_empty} that, for all $\state \in \dom$, we have
        \begin{align*}
            \min_{u \in \cset} \frac{d\clvf }{d \state } \cdot \cdyn (\state) \ctrl  &= -\gamma \clvf - \frac{d\clvf}{d \state } \cdot (\dyn (\state) +\ddyn(\state) \dstb^* ), \\
            & < -\hat \gamma \clvf - \frac{d\clvf}{d \state } \cdot (\dyn (\state) +\ddyn(\state) \dstb^* ).
        \end{align*}
    \end{proof}
\end{proposition}
Note here we allow $\hat \gamma$ to be $0$. By doing this, we still have 
\begin{align*}
     \min_{u \in \cset} \frac{d\clvf }{d \state } \cdot \cdyn (\state) \ctrl  < - \frac{d\clvf}{d \state } \cdot \dyn (\state).
\end{align*}
In other words, $\hat \gamma = 0$ guarantees \textbf{asymptotic} stabilizability. 

\begin{remark}
    Whether the inequality is strict in~\eqref{eqn:SCS_CLVF} is not critical. The inequality in~\eqref{eqn:SCS_CLVF} can be defined non-strict, because $\dot V_\gamma^\infty \leq -\gamma \clvf < 0$, and exponential stabilizability can be guaranteed, as is used  in~\cite{gong2024synthesizing}. The drawback is that in this case, choosing $\hat \gamma = 0$ cannot guarantee stabilizability (because the constraint will be $\dot V \leq 0$). Using strict inequality allows $\hat \gamma = 0$, and the convergence will be asymptotic. 
    % Note that to be consistent with the SCS given in~\eqref{eqn:StabilizeACS}, the inequality in~\eqref{eqn:SCS_CLVF} is strict, whereas in~\cite{gong2024synthesizing}, the inequality is not strict. As a result, the SCS is non-empty for all $\state \in \dom$ in~\cite{gong2024synthesizing}, but can be empty in~\eqref{eqn:SCS_CLVF}. 
\end{remark}

\begin{figure}[t]
    \centering
    \includegraphics[width=\columnwidth]{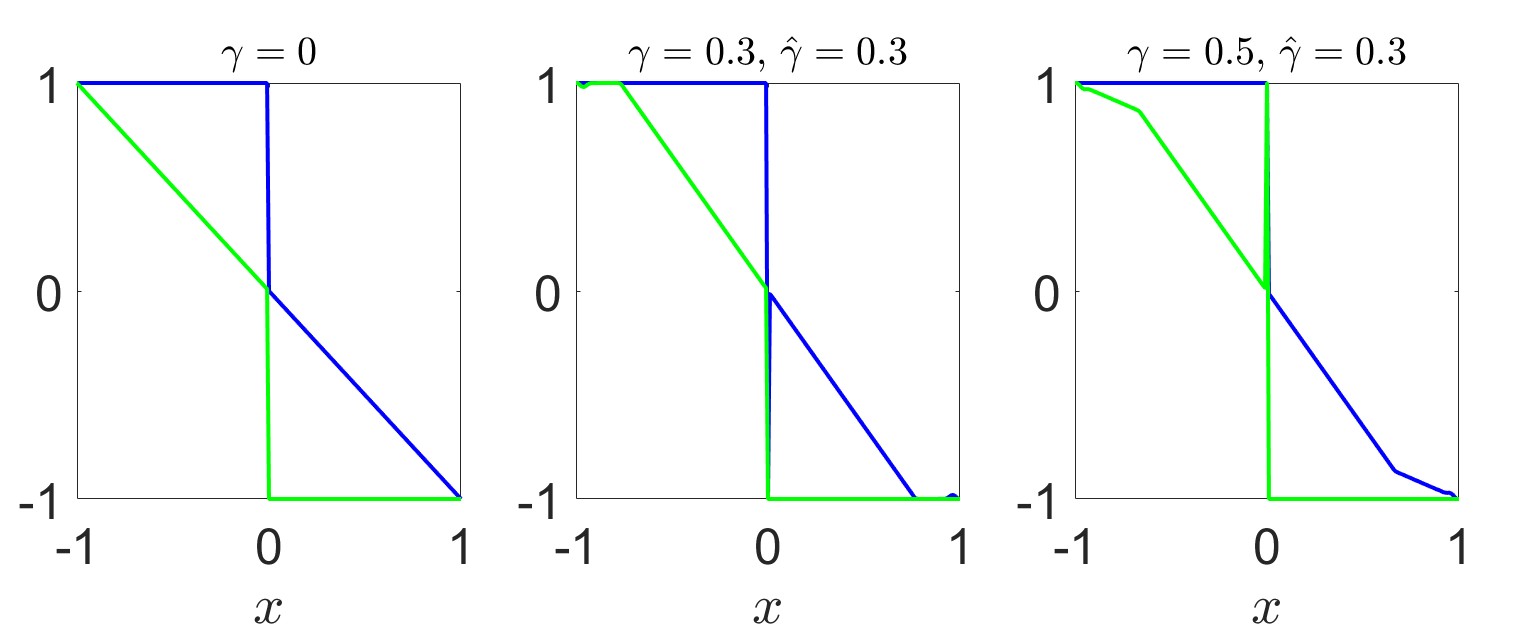}
    \vspace{-1em}
    \caption{The ACSs that guarantee asymptotically stabilizability (left) and exponential stabilizability with rate 0.3 (middle and right). The blue and green lines denote the upper and lower bounds for ACS, respectively. Since the inequality in~\eqref{eqn:SCS_CLVF2} is strict, the control cannot take the value on the bounds. When $\hat \gamma = \gamma$, there are some states whose upper and lower bounds of the ACS are the same (middle plot, near -1 and 1), and by definition, the ASC is empty for those states. However, this does not mean there exists no control that stabilizes the system with $\hat \gamma$ rate. As shown in the right, when picking $0.3 = \hat \gamma < \gamma = 0.5$, the ACS is non-empty for all $\state \in (-1,1)$. Though we cannot prove that $\sacs (\state ; \gamma_1, \hat \gamma) \subset  \sacs (\state ; \gamma_2, \hat \gamma)$ for $\gamma_1 < \gamma_2$, this example implies that when computing the R-CLVF, using a larger $\gamma$ will allow more flexibility for the ACS.}
    \vspace{-1em}
    \label{fig:diff_ACS}
\end{figure}

\textit{Running Example: 1D Cart. \\
Consider system given by $\dot \state = \state + \ctrl$, where $\ctrl \in [ -1, 1]$. It can be verified that for all $\state \in [-1,1]$, there exists a control s.t. the trajectory stays in $[-1,1]$. For all $\state \in (-1,1)$, there exists a control s.t. the trajectory can be stabilized to the origin \textbf{exponentially}, because there exists a finite time so that the trajectory reaches $x=0$. Further, the ACS that guarantees \textbf{asymptotic} stabilizability is given by $\sacs (\state) = {\ctrl \in (-\state, 1] \text{ if } \state < 0 \text{ and } } [-1, \state) \text{ if } \state > 0 $. The ACSs with different $\gamma$ and $\hat \gamma$ are shown in Fig.~\ref{fig:diff_ACS}.}

\subsection{Reach and Avoid ACS}
One sufficient condition for the trajectory to complete the reach or avoid task is that for every $\tinit$ and a small enough $\delta >0$, $\xi(\tinit+\delta;\tinit,\state,\ctrl)$ enters the zero sublevel set of $V(\state,\tinit+\delta)$. Formally, we have
% To be more specific, we set the following \textbf{value decay constraint}
\begin{align}\label{eqn:Value_constraint} 
    \dot  V(x,\tinit) = \frac{d\Vr}{d\tinit} +  \frac{d\Vr}{d\state} \cdot(\dyn(\state)+\cdyn(\state) \ctrl +\ddyn(\state) \dstb )  \leq 0.
\end{align}

Since both the value function and the trajectory are guaranteed to be continuous in $(\state,\tvar)$, for any $\state$ with $\Vr(\state,\tinit)<0$ and \textbf{any} $\ctrl \in \cset$, there is always a small enough time step $\delta$ s.t. the trajectory $\xi(\tinit+\delta;\tinit,\state,\ctrl)$ enters the zero sublevel set of $\Vr(\state,\tinit+\delta)$, i.e., $\dot \Vr(x,\tinit) \leq 0$. In other words, the ACS for $\Vr(\state,\tinit)<0$ is just $\cset$.

When $\Vr(\state,\tinit) = 0$, any control input satisfying~\eqref{eqn:Value_constraint} will result in the trajectory entering the zero sublevel set of $V(\state,\tinit+\delta)$, and therefore is an element of ACS. 

When $\Vr(\state,\tinit) > 0$, there exists no ACS. In summary, the reach ACS $\racs : \mathbb R^n \times [\tinit,\thor] \mapsto \cset$ is given by 
\begin{align} \label{eqn:ReachACS}
    \racs(\state,\tvar) = \{ \ctrl \in \cset : a(\state,\tvar) \ctrl & \leq b(\state,\tvar), \notag \\ &\text{ for all }\Vr(\state,\tvar) = 0  \},
\end{align}
where
\begin{align*}
    a = \frac{d\Vr}{d\state} \cdot \cdyn(\state), \hspace{.5em}  b=-\frac{d\Vr}{d\tinit} - \frac{d\Vr}{d\state} \cdot( \dyn(\state) + \ddyn (\state) \dstb^*).
\end{align*}

Following the same procedure, define the avoid ACS $\aacs : \mathbb R^n \times [\tinit,\thor] \mapsto \cset$ is given by 
\begin{align} \label{eqn: wtAvoidACS_finite}
    \aacs(\state,\tvar) = \{ \ctrl \in \cset : a(\state,\tvar) \ctrl & \leq b(\state,\tvar) \notag \\
    &\text{ for all }\Va(\state,\tvar) = 0 \},
\end{align}
with $a$ and $b$ in the same form as in the reach case. For some systems and obstacles, the avoid value function may converge to a fixed point. In that case, the value function is given by $\Va^\infty(\state) = \lim_{\tinit \rightarrow -\infty} \Va(\state,\tinit)$. Each $\alpha$-sublevel set of $\Va^\infty$ defines a robust control invariant set, and the zero-sublevel set is the \textbf{maximal} robust control invariant set of $\obs$. The corresponding ACS $\aacsinf: \mathbb R^n  \mapsto \cset$ is given by 
\begin{align} \label{eqn:AvoidACS_infinite}
    \aacsinf(\state) := \{ \ctrl \in \cset : a(\state) \ctrl \leq b(\state) , \text{ for all }\Va^\infty(\state) = 0 \},
\end{align}
with 
\begin{align*}
    a = \frac{d\Va^\infty}{d\state} \cdot \cdyn(\state) , \hspace{.5em} b(\state)= - \frac{d\Va^\infty}{d\state} \cdot (\dyn(\state)+\ddyn(\state) \dstb^*).
\end{align*}
%%%%%%%%%%%%%%%%%%%%%%%%%%%%%%%%%%%%%%%%%%%%%%%%%%%%%%%%%%%%%%%%%%%%%%%%%%%%%%%%%%%%%%%%%%%%%%%%%%%%%%%%%%%%%%%%%%%%%%%%%%%%%%%%%%%%%%%%%%%%%%%%%%%%%%%%%%%%
\section{Control Signal Synthesis}\label{SEC4} 
In this section, we show how to use the ACS to generate control signals so that the resulting trajectory completes the task (reach target/ avoid obstacle/ stabilize to $\mrcis$). 

Since we can get the optimal control input from~\eqref{eqn:HJoptCtrl}, one natural way is to treat the optimal control input as a feedback law $k \colon \mathbb R^n \mapsto \acs$, and close the loop. More generally, we could close the loop with any feedback laws whose image is the ACS. Then, the resulting closed-loop system should be able to complete the tasks, and once we have the trajectory, the control signal is simply $k(\traj)$.

% design the control signal is to let $\ctrl (\tvar) = \arg \min_{u}\frac{d V}{d\state}(\xi(\tvar;\tinit,\xinit,\ctrl_{[\tinit,\tvar)}(\cdot), \dstb{[\tinit,\tvar]}(\csig)),\tvar) \cdot \cdyn (x)$. In other words, the ``optimal'' control signal is concatenated by the optimal control input. As we have ACS, a more general way is to let $\ctrl (\tvar; \xinit) \in \acs \bigl(\xi(\tvar; \tinit, \xinit, \ctrl_{[\tinit,\tvar)} (\cdot),  \dstb_{[\tinit,\tvar)}(\cdot),\tvar)), \tvar \bigr)$, i.e., concatenating control inputs from ACS along the trajectory. 

However, this works only for continuous differentiable value functions, whereas all the value functions we mentioned are only Lipschitz continuous. Therefore, their gradients can be discontinuous (though the set of discontinuous states is measure zero). A direct result is that $a,b$ in the ACS can be discontinuous w.r.t $\state$, and at the discontinuous state, $a,b$ do not exist. This means a continuous feedback law $u = k(x)$, where $k \colon \mathbb R^n \mapsto \acs$, is not guaranteed to exist. That is to say, the classic solution for the closed-loop system 
\begin{align*}
    \dot \state = \dyn (\state) + \cdyn (\state)  k(\state) + \ddyn(\state) \cdot \dstb
\end{align*}
may fail to exist, not to mention the uniqueness~\cite{4518905}. 

To be more rigorous and also more consistent with the numerical implementation, we consider the sample-and-hold solution~\cite{clarke1997asymptotic}. Denote $\pi = \{ t_i\} _{i \geq 0}$ be a partition of $[\tinit, \infty)$, where $\tinit_i$ is a strictly increasing sequence and $\tinit_0 = \tinit$, $\tinit_i \rightarrow \infty$ as $i \rightarrow \infty$. The diameter of $\pi$ is defined as $\text{diam}(\pi) = \sup_i (\tinit_{i+1}- \tinit_i)$. Given any initial state $\state (\tinit) = \xinit $, feedback law $k(x)$ and partition $\pi$, the sample-and-hold trajectory (also called the $\pi-$trajectory) is given by recursively solving
\begin{align}\label{eqn:SAH-traj}
    \dot \state (\tvar) = \dyn (\state(\tvar)) + \cdyn (\state(\tvar)) k(\state (t_i)) + \ddyn(\state(\tvar))  \dstb,  
\end{align}
for $\tvar \in [\tinit_i,\tinit_{i+1}]$, and use the endpoint of the previous interval as the initial condition, for all $i = 0,1,2, ...$. Briefly speaking, within any interval, the control is constant. Compared with the classic solution, this $\pi-$trajectory is guaranteed to exist and be unique.

% solves the problem that the solution does not exist, which is caused by the Lipschitzness of value function: the possibility that the endpoint of a trajectory is non-differentiable is zero. 
% , and even if that does happen, we could very well use a smaller $t_{i+1}$. 

Now, we formally present the admissible control signal. Given any initial state with non-empty ACS, and a random feedback $k(x)$ whose image is the ACS, a partition $\pi$. Denote the resulting $\pi-$trajectory as $\xi_{\pi}(\tvar)$. The control signal $k(\xi_{\pi}(\tvar))$ is called an admissible control signal if $\reward(\xi_{\pi}(\thor)) \leq 0$ for the reach problem, $\loss(\xi_{\pi}(\tvar)) \leq 0$ for all $\tvar \in [\tinit, \thor]$ for the avoid problem, or $ \lim_{\tvar \rightarrow \infty } \min_{a \in \partial \mrcis(p)} \| \xi_{\pi}(\tvar)- a \| = 0  $ for the stabilize problem. In fact, the design of the feedback is unnecessary, and we could generate one control signal by the following process. 

Control signal synthesize process (CSSP): \textit{Given the initial state $\state$, partition $\pi$, randomly pick control $\ctrl_{0}$ from $\acs (\state, t_0) $, and solve~\eqref{eqn:dynsys} with $\csig = \ctrl_{0}$ for $[t_0,t_1]$. Denote the corresponding end state as $\state (t_1)$, and if the ACS at $\state (t_1)$ is non-empty, pick the control for the next interval from the ACS. If empty, pick from $\cset$. Repeat this process for $[t_i, t_{i+1}]$ for all $i>0$. }

With this, we could verify whether a control signal is admissible to the problem or not. This is important for us to verify the RAS set. For finite time reach and avoid problems, a control signal can be verified by the following proposition.

\begin{proposition} \label{prop:reach_avoid_condition}
    Given ACS~\eqref{eqn:ReachACS}~\eqref{eqn:AvoidACS_infinite} for all $\state$ and $\tvar \in [\tinit, \thor]$, and given partition $\pi = \{ t_i\} _{i = 0} ^ n$ s.t. $t_0 = t$, $t_n = 0$. Following the CSSP, if at each $t_i$, the ACS is non-empty, the resulting control signal is an admissible control signal. 
    
    \begin{proof}
        The proof is trivial. Since at each $\state (t_1)$, the ACS is non-empty, this mean means $V(\xi_{\pi} (t_i), t_i) \leq 0$ for all $i = 1,...,n$, therefore at $t_n = 0$, $V(\xi_{\pi} (t_n), 0)  \leq 0$. 
    \end{proof}
\end{proposition}

The admissible control signal of the infinite-time avoid problem can be similarly verified. For the stabilize problem, since the ACS is non-empty in $\dom$, the verification is simplified to check whether, at some point, the $\pi-$trajectory enters $\dom$. If so, we could use the following Lemma to show convergence to $\mrcis$. 
\begin{lemma} \label{lemma:R-CLVF-traj_converge}
    Given R-CLVF $\clvf$ and ACS $\sacs(\state; \hat \gamma, \gamma)$ with $0 \leq \hat \gamma < \gamma$, there exists $\delta > 0$ and $ 1-\hat \gamma < \alpha < 1$, s.t. for any partition $\pi = \{ t_i\} _{i \geq 0}$ with diameter $\delta$, $\clvf (\xi_{\pi}(t_i+1)) < \alpha \clvf (\xi_{\pi}(t_i)) $. 

    \begin{proof}
        From~\eqref{eqn:SCS_CLVF2}, we have
        \begin{align*}
            \frac{d\clvf }{d \state } \cdot \cdyn (\xi_{\pi}(t_i)) \ctrl  < &-\hat \gamma \clvf (\xi_{\pi}(t_i)) \\
            &- \frac{d\clvf}{d \state } \cdot (\dyn (\xi_{\pi}(t_i)) + \ddyn(\xi_{\pi}(t_i))\dstb^* ). 
        \end{align*}
        Because of the continuity of the $\pi-$trajectory, for all $\xi_{\pi}(t_i) \in \dom$ there exists $ \delta > 0$, $\hat \gamma < \bar \gamma < \gamma$ s.t. 
        \begin{align*}
            \clvf (\xi_{\pi}(t_i + \delta )) - \clvf (\xi_{\pi}(t_i  )) < \bar \gamma \clvf (\xi_{\pi}(t_i)),  
        \end{align*}
        and the proof is completed by letting $\alpha = 1-\bar \gamma > 1- \hat \gamma$.
    \end{proof}
\end{lemma}
In other words, numerically speaking, once the $\pi-$trajectory enters $\dom$, we could always find a small enough $\delta$ so that that trajectory then converges to $\mrcis$.

\section{Finding RAS Set with $\cset (\state) $} \label{SEC5} 
In this section, we show how the RAS set can be under-approximated by CSSP using the ACSs found in Section~\ref{SEC3}. The idea is to uniformly sample the state space, apply the CSSP and check if the trajectory can reach $\{\goal_i\}_{i = 1}^{q_1}$, avoid $\{\obs_i\}_{i = 1}^{q_2}$ and $\obs$ in the prespecified time horizon, and stabilize to $p$ (or $\mrcis$). With Proposition~\ref{prop:reach_avoid_condition}, for the reach and avoid task, we just need to check if the ACS is non-empty during the CSSP. For the stabilize problem, we derive some other conditions that will be explained later. 

% As mentioned before, one main benefit of using the ACS based approach is that the reach-avoid-stabilize problems can each have their own time horizons. Therefore, the methodology presented in this section can be easily extended to solving reach (and/or avoid) multiple targets (and/or obstacles), and ultimately converge to the POI. The original RAS problem is solved in the first subsection, while the more complexed problem with multiple targets and obstacles will be defined and solved in the second subsection. Further, since the idea of ACS actually originated from the self-contained subsystem decomposition, we could easily extend our method for finding RAS to relatively high-dimensional systems. 

% \subsection{Find RAS Set with Single Target and Obstacle}
Let us recall the RAS problem we aim to solve from Def~\ref{Def: RAS_set}. To find the ACSs, we first find the value function for reach, avoid, and stabilize separately using~\eqref{eqn:reach_value},~\eqref{eqn:avoid_value} and~\eqref{eqn:CLVF}. More specifically, the reach value functions are computed as $V_{r,i} ( \state, \tvar)$, $\tvar \in [ t_{r,i-1}-t_{r,i},0]$ for all $i = 1:q_1$ and the avoid value functions are computed for $V_{a,j} (\state, \tvar)$, $\tvar \in [ t_{a,j-1}-t_{a,j},0]$ for all $j = 1:q_2$. The reason is that for time-invariant dynamics~\eqref{eqn:dynsys}, reaching target starting from $\tinit $ until $ \treach$ is equivalent to reaching target starting from $\tinit -\treach  $ until $ \thor$. The same applies to the finite-time avoid problems too. For the infinite time avoid problem, letting $\tavoid \rightarrow \infty$ and compute the converged value function. The R-CLVF is also computed. For time-varying reach and avoid problems, the ACSs are only defined for $\tvar \in [\tinit , \treach]$ (and $\tvar \in [\tinit , \tavoid]$). When $t_{r,q_1} > t_{a,q_2}$, the avoid ACS is not defined for $[t_{a,q_2} , t_{r,q_1}]$, and is treated as $\cset$. In this case, lines 7 and 20 can be skipped. The same is applied when $t_{r,q1} < t_{a,q_2}$.

\begin{algorithm} [t]
\caption{Under-approximate RAS for Multiple Targets and Obastacles}\label{algo:Multi_ACS}
\begin{algorithmic}[1]
\Require Time series for finite-time reach and avoid problems $T_{r}, T_{a}$, and their ACSs $\mathcal {U}^{r,i}_{\mathcal G_i} (\state, \treach)$ and $\mathcal {U}^{a,i}_{\mathcal C_i} (\state, \tavoid)$. Infinite-time avoid ACS $\aacs (\state)$, stabilze $\sacs$, time step $\tstep$, and grid points $\{ x_i \}_{i = 1}^N $ \\

\textbf{Initialization:} $\tvar \gets t_{r,0}$, Index $\gets [1:N]$ \\
\textbf{Output:} states in $\{ x_i \}_{i = 1}^N $ that belongs to the RAS set 

\For{all grid points $\{ x_i \}_{i = 1}^N $}

\State \textit{Reach-Aviod Block: }

\While{$\tvar \leq  t_{r,q_1}$ }
\State $j \gets \min j $ s.t. $t_{r,j} \in T_r$, $\tvar \leq t_{r,j}$
\State $k \gets \min k $ s.t. $t_{a,k} \in T_a$, $\tvar \leq t_{a,k}$
\State $\mathcal U_{temp}( \state ) \gets \mathcal {U}^{r,j}_{\mathcal G_j} (\state,\tvar - t_{r,j}) \cap \mathcal {U}^{a,k}_{\mathcal O_k}  (\state,\tvar - t_{a,k}) \cap \aacs (\state) $

% \If{$\tvar \leq \tavoid$} 
% \State $\mathcal U_{temp}( \state ,\tvar) \gets \racs(\state,\tvar) \cap \aacs(\state,\tvar)$
% \Else \State $\mathcal U_{temp}( \state ,\tvar) \gets \racs(\state,\tvar) $
% \EndIf

\If {$\mathcal U_{temp}( \state ,\tvar) $ empty} 
\State Remove $i $ from Index 
\Else 
\State Sample $u$ from $\mathcal U_{temp}( \state ,\tvar)$ 
\State Get optimal disturbance $d^*$ from~\eqref{eqn:HJoptDstb}
\State Solve~\eqref{eqn:SAH-traj} and update $\state$
\State $\tvar \gets \tvar + \tstep$
\EndIf
\EndWhile

\State \textit{Stabilize-Aviod Block: }

\While{$x \notin \maxlevel$ }
\State $k \gets \min k $ s.t. $t_{a,k} \in T_a$, $\tvar \leq t_{a,k}$
\State $\mathcal U_{temp}( \state ) \gets \mathcal {U}^{a,k}_{\mathcal O_k}  (\state,\tvar - t_{a,k})\cap \sacs(\state) \cap \aacs (\state) $
\State do lines 9 to 16 
% \If {$\mathcal U_{temp}( \state ,\tvar) $ empty} 
% \State Remove $i $ from Index 
% \Else 
% \State Sample $u$ from $\mathcal U_{temp}( \state ,\tvar)$ 
% \State Get optimal disturbance $d^*$ from~\eqref{}
% \State $x \gets x + (\dyn(x)+\cdyn(x)u + \ddyn (x)d^*)\tstep$
% \State $\tvar \gets \tvar + \tstep$
% \EndIf

\EndWhile

\EndFor

\end{algorithmic}
\end{algorithm}

After computing the value functions, the ACSs are obtained by~\eqref{eqn:ReachACS} and~\eqref{eqn: wtAvoidACS_finite}, and denoted as $\mathcal {U}^{r,i}_{\mathcal G_i}$ and $\mathcal {U}^{a,i}_{\mathcal C_i}$. The ACS for infinite-time avoid problem and R-CLVF are computed by~\eqref{eqn:SCS_CLVF2} and~\eqref{eqn:AvoidACS_infinite}, and are denoted as $\aacs, \sacs$. These ACSs (more specially the $a,b$'s), together with diameter $\delta$, time series $T_r,T_a$, and N grid points $\{ x_i \}_{i=1}^N$ are the input of Alg.~\ref{algo:Multi_ACS}. 
The output is the \textbf{Index}, i.e., the index $i$ of the initial states that belong to the RAS set. 

There are two main blocks in Alg.~\ref{algo:Multi_ACS}, the reach-avoid block and the stabilize-avoid block. In both blocks, at each time step, we first find the corresponding target we want to reach and the obstacle we want to avoid (Lines 6, 7, 20). We then find the ACSs for reach-avoid and stabilize-avoid by taking the intersection of the corresponding reach and avoid (stabilize) ACSs $\mathcal {U}^{a,i}_{\mathcal C_i}$ and $\mathcal {U}^{r,i}_{\mathcal G_i}$ (or $\sacs$) and $\aacs$ (Lines 8, 21). If the intersection is not empty, we sample a random control input and propagate the trajectory. If it's empty, the index of the initial state is removed from \textbf{Index}. Note that for practical implementation, we could put all grid points into a big $n-by-N$ matrix, and do all the computation on this big matrix. Alg.~\ref{algo:Multi_ACS} is written in terms of ``for all grid points'' just to make the algorithm easier to understand.

% for $\treach-\tinit$ (and $\tavoid - \tinit$) is that in HJR, the terminal time is fixed and the value function is computed backwards from $\thor$ to a specified time $\tinit$, while here, we have to fix the initial time as $\tinit$ and check . This is allowed because the dynamics~\eqref{eqn:dynsys} is time-invariant.  

For finite-time reach and avoid problems, it is easy to set up the end-loop condition. However, the stabilization problem relates to an infinite time horizon. Simply setting a very large time horizon increases computation time and is unnecessary. To solve it, we use Lemma~\ref{lemma:R-CLVF-traj_converge}, and the end-loop condition for the stabilization is set as whether the trajectory enters the set $\maxlevel$, where
$$\maxlevel \coloneqq \{ \max a \text{ s.t. } \forall \state \text{ with } \clvf (\state) \leq a, \loss (\state) \leq 0 \}$$
is the largest sub-level set of $\clvf$ that is not intersected with $\obs$. Once the trajectory enters $\maxlevel$, the trajectory will decay to a low-level set and ultimately converge to $\mrcis$. In addition, safety is guaranteed because the trajectory can stay in $\maxlevel$ for all time. The $\hat \gamma$ we use in~\eqref{eqn:SCS_CLVF2} determines the convergence rate and the RAS set. 

% Therefore, this is stronger than $\lim_{\tvar \rightarrow \infty } \min_{a \in \partial \mrcis} \| \traj - a \| = 0$, and $\traj \notin \obs$. When in~\eqref{eqn:SCS_CLVF2}, $\hat \gamma = 0$, the conditions are equivalent. The algorithm for finding the RAS set is shown in Alg.~\ref{algo:finding_ACS}.

\begin{figure}[t]
    \centering
    \includegraphics[width=\columnwidth]{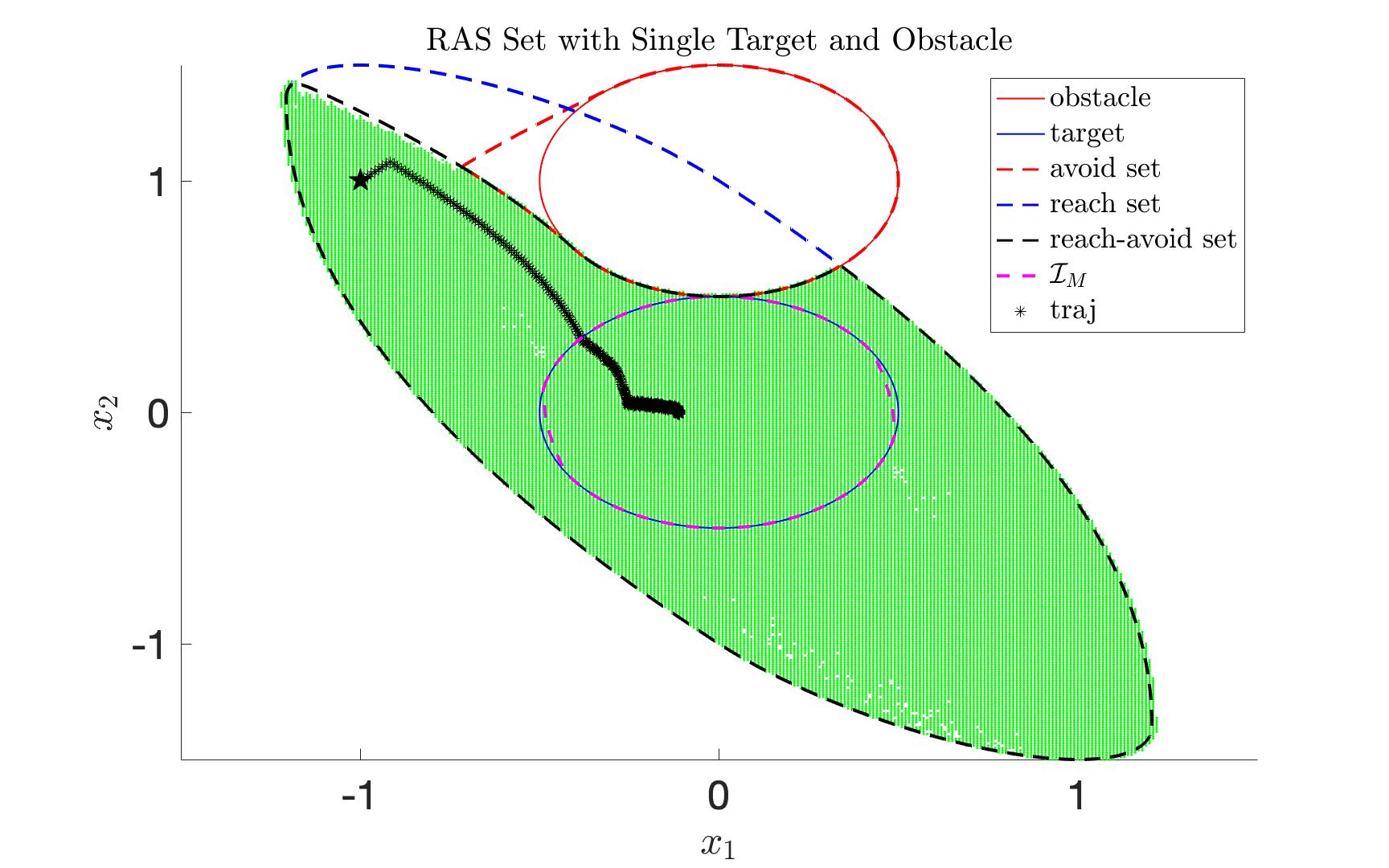}
    \vspace{-1em}
    \caption{The RAS set from Alg.~\ref{algo:Multi_ACS}, shown in green. The target (obstacle) and reach (avoid) sets are shown in the blue (red) line and blue (red) dashed line, and the reach-avoid set from HJR is shown in blake dashed line. One trajectory with initial state $[-1;1]$ (black pentagram) is shown in black stars. The location of the obstacle and the dynamics of the system make it impossible for the system to hit the obstacle after reaching the target in $[0,1]$ and before entering $\maxlevel$. The total iteration is 144 and takes 39.13s with 401*401 initial states.}
    \vspace{-1em}
    \label{fig:singleRAS}
\end{figure}

\begin{theorem} \label{thrm:RAS}
    Alg.~\ref{algo:Multi_ACS} under-approximates the RAS set. 
    
    \begin{proof}
        Given $\state_i$, if $i \in $ \textbf{Index}, this means in the Reach-Avoid Block, for all $\tvar \in [t_{r,i}, t_{r,i+1}]$, $i = 1,...,q_1$, and for all $\tvar \in [t_{a,j}, t_{a,j+1}]$, $j = 1,...,q_1$, the trajectory satisfies $\loss_j (\xi_{\pi}(\tvar)) \leq 0$ and $\reward_i (\xi_{\pi}(t_{r,i})) \leq 0$. Further, in the Stabilize-Avoid Block, for all $\tvar \in [t_{r,q_1}, t_{a,q_2}]$, the trajectory satisfies $\loss_{q_2} (\xi_{\pi}(\tvar)) \leq 0$ and $\xi_{\pi}(\tau) \in \maxlevel$ for some $\tau$. If the obstacle is to be avoided for all time, then it must be $\loss (\xi_{\pi}(\tvar)) \leq 0$ for all $\tvar \in [\tinit, \tau]$. Combined with Lemma~\ref{lemma:R-CLVF-traj_converge}, this trajectory satisfies the RAS requirement. The control signal that renders the system complete the RAS task can be obtained following the CSSP, by plugging the sampled control input from Alg.~\ref{algo:Multi_ACS}. 

        On the other hand, unfortunately, if a state belongs to the RAS set, Alg.~\ref{algo:Multi_ACS} may characterize it as \textbf{not} in the RAS set. This is because Alg.~\ref{algo:Multi_ACS} do not look ahead in time: it will not try to avoid any obstacle until it reaches the zero level set of $\Va(x,t) $, and when it's close to the zero level set of $\Va(x,t) $, it already wasted too much time so that it cannot reach the target in the prescribed time. 
    \end{proof}
\end{theorem}

% The input of Alg.~\ref{algo:finding_ACS} are the ACSs for the reach problem, avoid problem, and stabilize problem, a small enough time step $\tstep$, and N grid points $\{ x_i \}_{i=1}^N$. When the obstacle is to be avoided for all time, use $\aacsinf$ instead of $\aacs$. The output is the index $ind$ of $\{ x_i \}_{i=1}^N$ that belongs to the RAS set. 

\begin{remark} \label{remark:reach_bdry}
   In the Reach-Avoid Block of Alg.~\ref{algo:Multi_ACS}, when $V_{r,i}(\state,\tvar) < 0$ for some state $\state$, the reach ACS is not active, and the trajectory may travel to anywhere in next time step with an arbitrary control input. It only starts to reach the goal once $V_{r,i} (\state,\tvar) = 0$. As a result, for all states that do not start inside the target, they will reach the boundary of the target at the same time $\treach$.
\end{remark}

\section{Numerical Examples} \label{SEC6} 
\begin{figure}[t]
    \centering
    \includegraphics[width=\columnwidth]{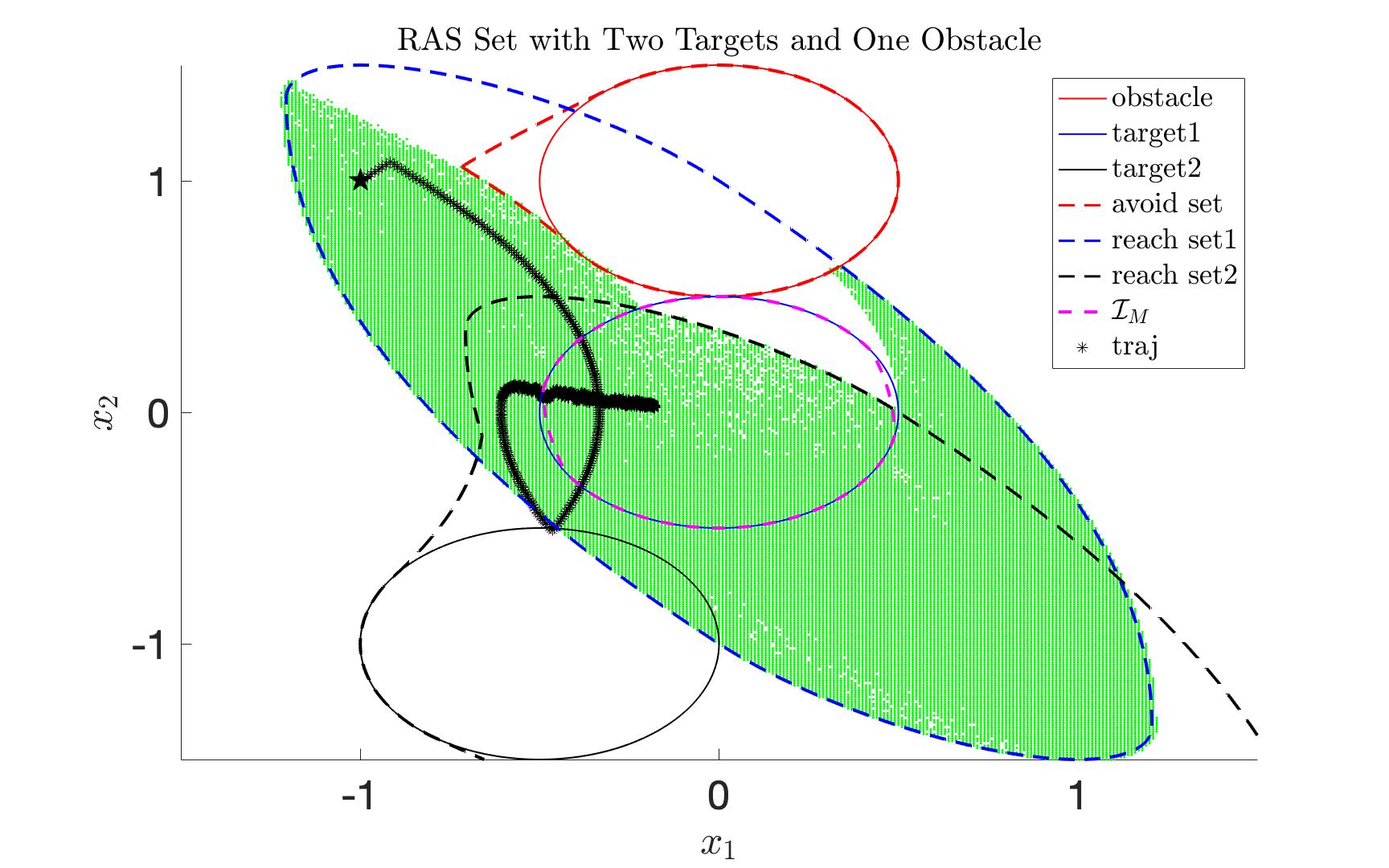}
    \vspace{-1em}
    \caption{The RAS set from Alg.~\ref{algo:Multi_ACS}, shown in green. Target 1 (target 2, obstacle) and reach set1 (reach set2, avoid set) are shown in the blue (black, red) lines and blue (black, red) dashed lines. One trajectory with initial state $[-1;1]$ (black pentagram) is shown in black stars. From Remark~\ref{remark:reach_bdry}, at $s = t_{r,1}$, all the states will reach exactly the boundary of target 1. Therefore, the states that can then reach target 2 are the states that reach the intersection of reach set2 and target 1 at $s = t_{r,1}$. The total iteration is 580 and takes 149.40s with 401*401 initial states.}
    \vspace{-1em}
    \label{fig:multiRAS}
\end{figure}

% \zgnote{please also check 'main\_version2.tex'. In current file, I only provide double integrator. In version2, I delete fig 2 and add the dubins car example.}

We provide two simulations to validate the theory. The simulation is based on MATLAB and solved by helperOC and ToolboxLS. Detailed code can be found \href{https://github.com/ZG0327/CDC2025_RAS.git}{https://github.com/ZG0327/CDC2025\_RAS.git}.
\subsection{Double Intergrator}
\begin{figure}[t]
    \centering
    \includegraphics[width=\columnwidth]{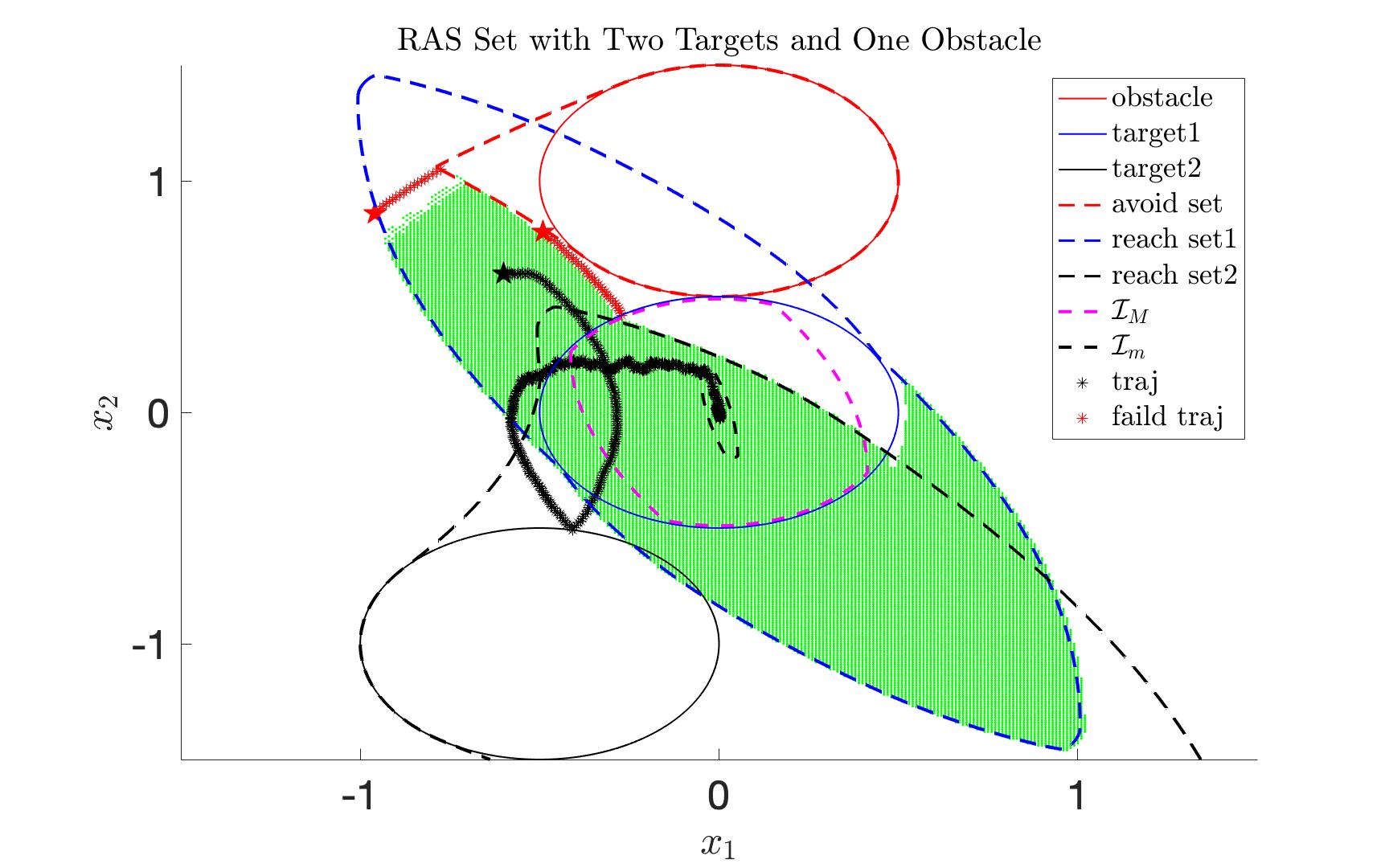}
    \vspace{-1em}
    \caption{The RAS set from Alg.~\ref{algo:Multi_ACS}, shown in green. Target 1 (target 2, obstacle) and reach set1 (reach set2, avoid set) are shown in the blue (black, red) lines and blue (black, red) dashed lines. One successful trajectory with initial state $[-0.6;0.6]$ (black pentagram) is shown in black stars, and two failed trajectories with initial states $[-0.96,0.86],[-0.49;0.78]$ (red pentagram) are shown in red stars. The failed trajectory either hits the avoid set or does not enter the intersection of the reach set 2 and target 1. The total iteration is 235 and takes 153.34s with 401*401 initial states.}
    % \vspace{-1em}
    \label{fig:multiRAS_d}
\end{figure}

The system dynamics is given by
\begin{align*}
    \dot x_1 = x_2 + d, \hspace{.5em} \dot x_2 = u, 
\end{align*}
where $u \in [-1,1]$. We consider three problems: 
\begin{enumerate}
    \item One target and obstacle, $d = 0$: reach $\goal = \{ x : \|x\|_2  \leq 0.5 \}$ within $[0,1]$, then stabilize to the origin, and avoid obstacle $\obs = \{x : \|x - [0;1] \| \leq 0.5  \}$ for all time. 

    \item Two targets and one obstacle, $d = 0$: reach $\goal_1 = \{ x : \|x\|_2\leq 0.5 \}$ within $\tvar \in [0,1]$, then reach $\goal_2 = \{ x : \|x - [-0.5;-1]\|_2 \leq 0.5 \}$ within $\tvar \in[1,2]$ (i.e., $T_r = [0,1,2]$), then stabilize to the origin, and avoid obstacle $\obs = \{x : \|x - [0;1] \| \leq 0.5 \}$ for all time. 

    \item Same targets, obstacle, and POI as 2), $d \in [-0.2,0.2]$.

\end{enumerate}
The results are shown in Fig.~\ref{fig:singleRAS}, Fig.~\ref{fig:multiRAS} and Fig.~\ref{fig:multiRAS_d}, respectively. In problem 1, Alg.~\ref{algo:Multi_ACS} provides almost identical results compared to directly solving the reach-avoid value function. This is because $\maxlevel$ is almost the same as the target; therefore, once the system reaches the target, it just takes a few time steps for it to enter $\maxlevel$. In problem 3, the POI is no longer a stable equilibrium point, and the SRCIS is shown as the black dashed line. The trajectory starting from [-0.96;0.86] fails because the ACS does not consider the long-term effect, i.e., it is only point-wise admissible. This is also why Alg.~\ref{algo:Multi_ACS} is only an under-approximation. 

\subsection{3D Dubins Car}
\begin{figure}[t]
    \centering
    \includegraphics[width=\columnwidth]{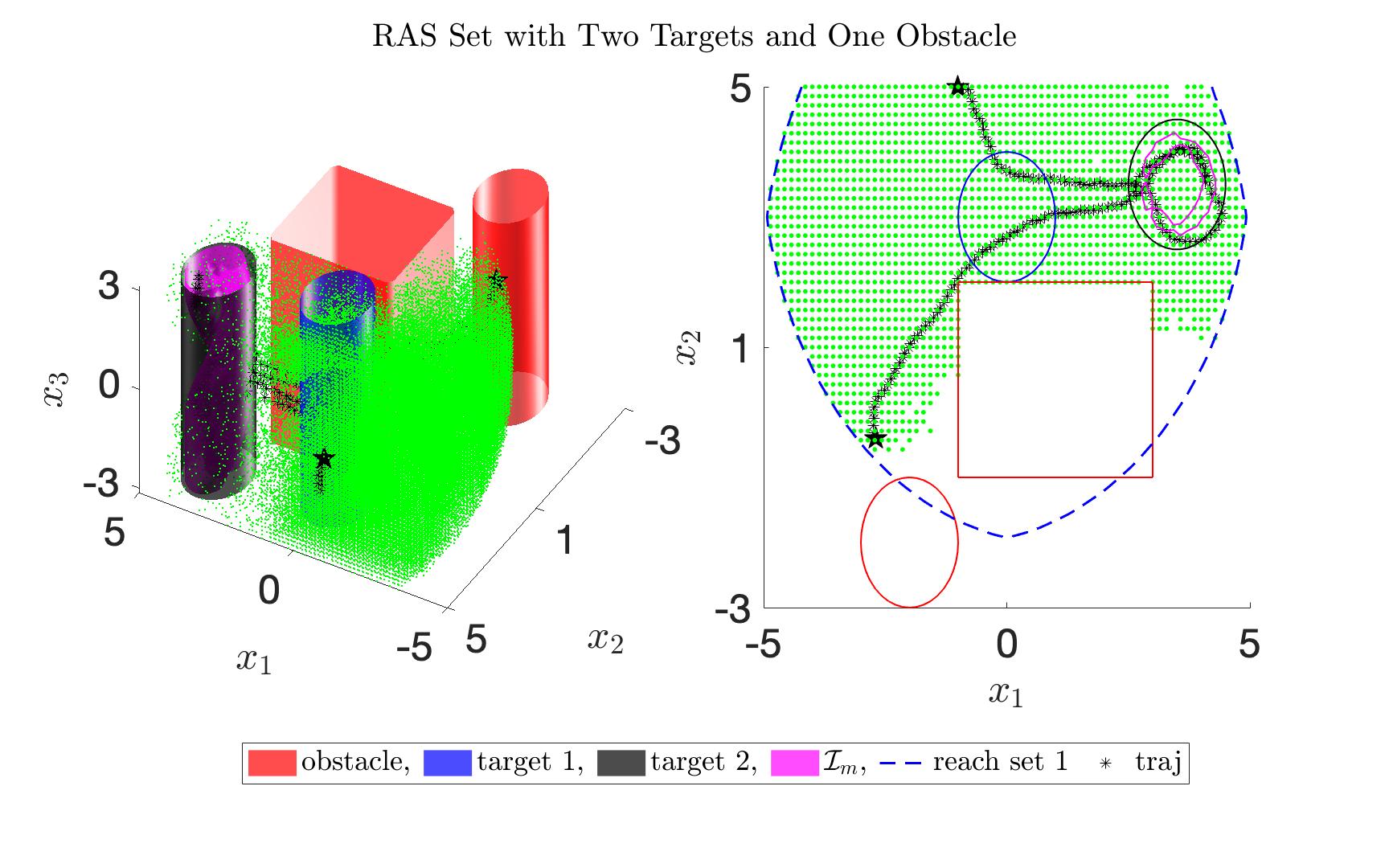}
    \vspace{-1em}
    \caption{The RAS set from Alg.~\ref{algo:Multi_ACS}, shown in green. Target 1, target 2, obstacle are shown in the blue, black, and red lines, and $\mrcis$ is shown in magenta. Two successful trajectories with initial states $[-1;5;0], [-2.7;-0.4;1.9]$ (black pentagram) are shown in black stars: they first reach target 1, then reach target 2, and finally stabilize to $\mrcis$. Left, the original 3D plot. Right, the projected plots in $x_1-x_2$ plane. The total iteration is 132 and takes 189.78s with 71*71*51 initial states.}
    \vspace{-1em}
    \label{fig:dubins}
\end{figure}

The system dynamics is given by 
\begin{align*}
    \dot x_1 = v \cos(x_3)+d_1, \hspace{1em}, \dot x_2 = v \sin(x_3)+d_2, \hspace{1em}\dot x_3 = u,
\end{align*}
where $d_1,d_2 \in [-0.2,0.2]$, $u\in[-1,1]$, $v = 1$. Assume we want to reach target 1 $\goal_1 = \{ x : x_1^2+(x_2-3)^2 \leq 1 \}$ in [0,5] second, reach target 2 $\goal_2 = \{ x : (x_1-3.5)^2+(x_2-3.5)^2 \leq 1 \}$ in [5,13] second, then stabilize to $p = [3.5;3.5; 0]$, while avoid the obstacle $\obs = \{ x: (x_1+2)^2+(x_2+2)^2 \leq 1, \text{ or } -1\leq x_1 \leq 3 \text{ and } -1\leq x_2 \leq 2, \}$. The result is shown in Fig.~\ref{fig:dubins}. In this example, target 1 is a subset of the reach set 2 (not visualized), and $p$ is not an equilibrium point.

\begin{comment}
    
Let's consider a 3D dubin's car example with constant velocity: 
\begin{align*}
    \dot x_1 = v \cos(x_3)+d_1, \hspace{1em}, \dot x_2 = v \sin(x_3)+d_2, \hspace{1em}\dot x_3 = u,
\end{align*}
where $d_1,d_2 \in [-0.2,0.2]$, $u\in[-1,1]$, $v = 1$. Assume we want to reach target 1 $\goal_1 = \{ x : x_1^2+(x_2-3)^2 \leq 1 \}$ in [0,5] second, reach target 2 $\goal_1 = \{ x : (x_1-3.5)^2+(x_2-3.5)^2 \leq 1 \}$ in [5,13] second, then stabilize to $p = [3.5;3.5; \text{any } x_3]$, while avoid the obstacle $\obs = \{ x: (x_1+2)^2+(x_2+2^2 \leq 1, \hspace{.5em} -1\leq x_1 \leq 3, \hspace{.5em} -1\leq x_2 \leq 2, \}$. The result is shown in~\ref{fig:dubins}. 
\begin{figure}[t]
    \centering
    \includegraphics[width=\columnwidth]{figures/fig_dubins.jpg}
    \caption{The RAS set from Alg.~\ref{algo:Multi_ACS}, shown in green. Target 1, target 2, obstacle are shown in the blue, black, red lines, and $\mrcis$ is shown in magenta. Two successful trajectories with initial state $[-1;5;0], [-2.7;-0.4;1.9]$ are shown in black stars: it first reaches target 1, then reaches target2, and finally stabilized to $\mrcis$. Left, the original 3D plot. Right, the projected plots in $x_1-x_2$ plane. The total iteration is 132 and takes 189.78s with 71*71*51 initial states.}
    \label{fig:dubins}
\end{figure}
\end{comment}

%%%%%%%%%%%%%%%%%%%%%%%%%%%%%%%%%%%%%%%%%%%%%%%%%%%%%%%%%%%%%%%%%%%%%%%%%%%%%%%%%%%%%%%%%%%%%%%%%%%%%%%%%%%%%%%%%%%%%%%%%%%%%%%%%%%%%%%%%%%%%%%%%%%%%%%%%%%%
\section{Conclusion}

In this paper, we propose a forward propagation method to find the RAS set. Compared with existing methods, the main benefits are: 1) the proposed method leverages existing theory from HJR and does not require defining new value functions; 2) the proposed algorithm solves for RAS problem with multiple targets and obstacles, that have to be reached or avoided in different time horizons specified; 3) The algorithm is proved to be an underapproximation, guaranteeing safety. 

Future directions include using learning methods to find ACS, computing the HJR value function with the ACSs for reach/avoid/stabilize problems, and combining learning methods to exactly recover the RAS set. 

% \vspace{2.5em}

\bibliographystyle{IEEEtran}%\begin{scriptsize}
\bibliography{ref}%\end{scriptsize}

\begin{thebibliography}{10}
\providecommand{\url}[1]{#1}
\csname url@samestyle\endcsname
\providecommand{\newblock}{\relax}
\providecommand{\bibinfo}[2]{#2}
\providecommand{\BIBentrySTDinterwordspacing}{\spaceskip=0pt\relax}
\providecommand{\BIBentryALTinterwordstretchfactor}{4}
\providecommand{\BIBentryALTinterwordspacing}{\spaceskip=\fontdimen2\font plus
\BIBentryALTinterwordstretchfactor\fontdimen3\font minus \fontdimen4\font\relax}
\providecommand{\BIBforeignlanguage}[2]{{%
\expandafter\ifx\csname l@#1\endcsname\relax
\typeout{** WARNING: IEEEtran.bst: No hyphenation pattern has been}%
\typeout{** loaded for the language `#1'. Using the pattern for}%
\typeout{** the default language instead.}%
\else
\language=\csname l@#1\endcsname
\fi
#2}}
\providecommand{\BIBdecl}{\relax}
\BIBdecl

\bibitem{fisac2015reach}
J.~F. Fisac, M.~Chen, C.~J. Tomlin, and S.~S. Sastry, ``Reach-avoid problems with time-varying dynamics, targets and constraints,'' in \emph{Proceedings of the 18th international conference on hybrid systems: computation and control}, 2015, pp. 11--20.

\bibitem{bansal2017hamilton}
S.~Bansal, M.~Chen, S.~Herbert, and C.~J. Tomlin, ``Hamilton-jacobi reachability: A brief overview and recent advances,'' in \emph{2017 IEEE 56th Annual Conference on Decision and Control (CDC)}.\hskip 1em plus 0.5em minus 0.4em\relax IEEE, 2017, pp. 2242--2253.

\bibitem{freeman1996control}
R.~A. Freeman and J.~A. Primbs, ``Control lyapunov functions: New ideas from an old source,'' in \emph{Proceedings of 35th IEEE conference on decision and control}, vol.~4.\hskip 1em plus 0.5em minus 0.4em\relax IEEE, 1996, pp. 3926--3931.

\bibitem{sontag1999control}
E.~D. Sontag, ``Control-lyapunov functions,'' in \emph{Open problems in mathematical systems and control theory}.\hskip 1em plus 0.5em minus 0.4em\relax Springer, 1999, pp. 211--216.

\bibitem{primbs1999nonlinear}
J.~A. Primbs, V.~Nevisti{\'c}, and J.~C. Doyle, ``Nonlinear optimal control: A control lyapunov function and receding horizon perspective,'' \emph{Asian Journal of Control}, vol.~1, no.~1, pp. 14--24, 1999.

\bibitem{xu2015robustness}
X.~Xu, P.~Tabuada, J.~W. Grizzle, and A.~D. Ames, ``Robustness of control barrier functions for safety critical control,'' \emph{IFAC-PapersOnLine}, vol.~48, no.~27, pp. 54--61, 2015.

\bibitem{ames2016control}
A.~D. Ames, X.~Xu, J.~W. Grizzle, and P.~Tabuada, ``Control barrier function based quadratic programs for safety critical systems,'' \emph{IEEE Transactions on Automatic Control}, vol.~62, no.~8, pp. 3861--3876, 2016.

\bibitem{ames2019control}
A.~D. Ames, S.~Coogan, M.~Egerstedt, G.~Notomista, K.~Sreenath, and P.~Tabuada, ``Control barrier functions: Theory and applications,'' in \emph{2019 18th European control conference (ECC)}.\hskip 1em plus 0.5em minus 0.4em\relax Ieee, 2019, pp. 3420--3431.

\bibitem{garg2019control}
K.~Garg and D.~Panagou, ``Control-lyapunov and control-barrier functions based quadratic program for spatio-temporal specifications,'' in \emph{2019 IEEE 58th Conference on Decision and Control (CDC)}.\hskip 1em plus 0.5em minus 0.4em\relax IEEE, 2019, pp. 1422--1429.

\bibitem{zeng2021safety}
J.~Zeng, B.~Zhang, and K.~Sreenath, ``Safety-critical model predictive control with discrete-time control barrier function,'' in \emph{2021 American Control Conference (ACC)}.\hskip 1em plus 0.5em minus 0.4em\relax IEEE, 2021, pp. 3882--3889.

\bibitem{romdlony2016stabilization}
M.~Z. Romdlony and B.~Jayawardhana, ``Stabilization with guaranteed safety using control lyapunov--barrier function,'' \emph{Automatica}, vol.~66, pp. 39--47, 2016.

\bibitem{braun2020comment}
P.~Braun and C.~M. Kellett, ``Comment on “stabilization with guaranteed safety using control lyapunov--barrier function”,'' \emph{Automatica}, vol. 122, p. 109225, 2020.

\bibitem{meng2023lyapunov}
Y.~Meng and J.~Liu, ``Lyapunov-barrier characterization of robust reach--avoid--stay specifications for hybrid systems,'' \emph{Nonlinear Analysis: Hybrid Systems}, vol.~49, p. 101340, 2023.

\bibitem{dawson2022safe}
C.~Dawson, Z.~Qin, S.~Gao, and C.~Fan, ``Safe nonlinear control using robust neural lyapunov-barrier functions,'' in \emph{Conference on Robot Learning}.\hskip 1em plus 0.5em minus 0.4em\relax PMLR, 2022, pp. 1724--1735.

\bibitem{chenevert2024}
\BIBentryALTinterwordspacing
G.~Chenevert, J.~Li, A.~kannan, S.~Bae, and D.~Lee, ``Solving reach-avoid-stay problems using deep deterministic policy gradients,'' 2024. [Online]. Available: \url{https://arxiv.org/abs/2410.02898}
\BIBentrySTDinterwordspacing

\bibitem{li2024certifiable}
\BIBentryALTinterwordspacing
J.~Li, D.~Lee, J.~Lee, K.~S. Dong, S.~Sojoudi, and C.~Tomlin, ``Certifiable reachability learning using a new lipschitz continuous value function,'' \emph{IEEE Robotics and Automation Letters}, vol.~9, no.~2, pp. 1--8, 2024. [Online]. Available: \url{https://arxiv.org/pdf/2408.07866}
\BIBentrySTDinterwordspacing

\bibitem{he2023efficient}
C.~He, Z.~Gong, M.~Chen, and S.~Herbert, ``Efficient and guaranteed hamilton-jacobi reachability via self-contained subsystem decomposition and admissible control sets,'' \emph{IEEE Control Systems Letters}, 2023.

\bibitem{gong2024synthesizing}
Z.~Gong, H.~J. Jeong, and S.~Herbert, ``Synthesizing control lyapunov-value functions for high-dimensional systems using system decomposition and admissible control sets,'' \emph{arXiv preprint arXiv:2404.01829}, 2024.

\bibitem{gong2024robust}
\BIBentryALTinterwordspacing
Z.~Gong and S.~Herbert, ``Robust control lyapunov-value functions for nonlinear disturbed systems,'' 2024. [Online]. Available: \url{https://arxiv.org/abs/2403.03455}
\BIBentrySTDinterwordspacing

\bibitem{4518905}
J.~Cortes, ``Discontinuous dynamical systems,'' \emph{IEEE Control Systems Magazine}, vol.~28, no.~3, pp. 36--73, 2008.

\bibitem{clarke1997asymptotic}
F.~H. Clarke, Y.~S. Ledyaev, E.~D. Sontag, and A.~I. Subbotin, ``Asymptotic controllability implies feedback stabilization,'' \emph{IEEE Transactions on Automatic Control}, vol.~42, no.~10, pp. 1394--1407, 1997.

\end{thebibliography}

\end{document}